\pdfoutput=1
\documentclass[manuscript,screen,nonacm]{acmart}

\acmJournal{CSUR}
\usepackage{amsmath}
\usepackage{booktabs}
\usepackage{multicol}
\usepackage{subcaption}
\usepackage{xkcdcolors}
\usepackage{tabularray}
\usepackage{microtype}
\usepackage{fontawesome5}
\usepackage{colortbl}
\usepackage{soul}
\usepackage{makecell}
\usepackage{enumitem}
\usepackage{tabularx}
\usepackage{anyfontsize}

\begin{document}

\title{Natural Language Generation from Visual Events: State-of-the-Art and Key Open Questions}

\author{Aditya K Surikuchi}
\email{a.k.surikuchi@uva.nl}
\author{Raquel Fern{\'a}ndez}
\email{raquel.fernandez@uva.nl}
\author{Sandro Pezzelle}
\email{s.pezzelle@uva.nl}
\affiliation{
  \institution{Institute for Logic, Language and Computation (ILLC), University of Amsterdam}
  \city{Amsterdam}
  \country{Netherlands}
}

\renewcommand{\shortauthors}{Surikuchi, Fern{\'a}ndez, Pezzelle}

\begin{abstract}
  In recent years, a substantial body of work in visually grounded natural language processing has focused on real-life multimodal scenarios such as describing content depicted in images or videos. However, comparatively less attention has been devoted to studying the nature and degree of interaction between the different modalities in these scenarios. In this paper, we argue that any task dealing with natural language generation from sequences of images or frames is an instance of the broader, more general problem of modeling the intricate relationships between visual events unfolding over time and the features of the language used to interpret, describe, or narrate them. Therefore, solving these tasks requires models to be capable of identifying and managing such intricacies. We consider five seemingly different tasks, which we argue are compelling instances of this broader multimodal problem. Subsequently, we survey representative modeling and evaluation approaches adopted for these tasks in recent years and examine the common set of challenges these tasks pose. Building on this perspective, we identify key open questions and propose several research directions for future investigation.
\end{abstract}



\maketitle
\section{Introduction}
\label{sec:1}

The ability to reason over visual events and describe them in natural language is a cornerstone of human communication and a crucial capability for intelligent systems operating in dynamic multimodal environments. From assistive technologies that interpret the world for visually impaired users to AI agents that must understand and narrate video content or interact with humans in real time, the demand for systems that can bridge visual perception and language over a temporal span is growing rapidly.

Initially, much of the progress in visually grounded natural language processing (NLP) focused on language generation from a static image. Tasks such as image captioning~\cite{vinyals2015show} and visual question answering~\cite{antol2015vqa} have demonstrated the potential of multimodal models to ground linguistic expressions in the visual modality~\cite{mm_reasoning0,mm_reasoning1,mm_reasoning2}. Yet, even a single image is rarely described in isolation. Human-written image descriptions often reflect the background knowledge and expectations of their authors, e.g., when mentioning what is not depicted~\cite{vanmiltenburg2016pragmatic}. Moreover, for accessibility purposes, visually impaired users prefer descriptions that account for the context in which the image appears, such as the surrounding webpage~\cite{mohanbabu2024context}. From this perspective, describing an image in context can be considered a special case of the problem we address in this paper, in which the visual sequence consists of a single image. Several of the challenges we discuss in Section~\ref{sec:challenges}, such as visual grounding and knowledge integration, are also relevant to image captioning. However, the world is not static. Real-world applications such as media content creation, real-time sports commentary and news reporting, and interactive multimodal assistants frequently require reasoning about events as they unfold over time. This requires capturing key aspects of event understanding, including temporal dynamics, causality, and narrative structure, that arise from complex multimodal interaction.

In this paper, we argue for a unified perspective: that all NLP tasks dealing with visual events, whether presented through videos or ordered image sets, are instances of a broader, more general problem. At the heart of this problem lies the need to model the intricate relationship between temporally ordered visual events and the structure, content, and function of the language used to interpret, describe, or narrate them. We claim that solving these tasks effectively requires models to be capable of identifying and managing such intricacies, leveraging the temporal, causal, and discourse-level patterns inherent in both visual data and language, as well as the many ways in which these aspects interact.

\begin{figure}[h]
  \centering
  \begin{tblr}{
      colspec = {X[c,h]X[c]},
      stretch = 0,
      rowsep  = 5pt,
      hline{1,3,5,7,9,11} = {1-Z}{0.35pt},
      vline{1,3} = {1-Z}{0.35pt}
    }
    <\textit{Describe the changes from the left image to the right image.}, \color{xkcdOlive}\faImages\color{black}> & \textit{Change Captioning} \\
    \vspace{-0.275cm}
    \includegraphics[width=\linewidth,height=1.75cm]{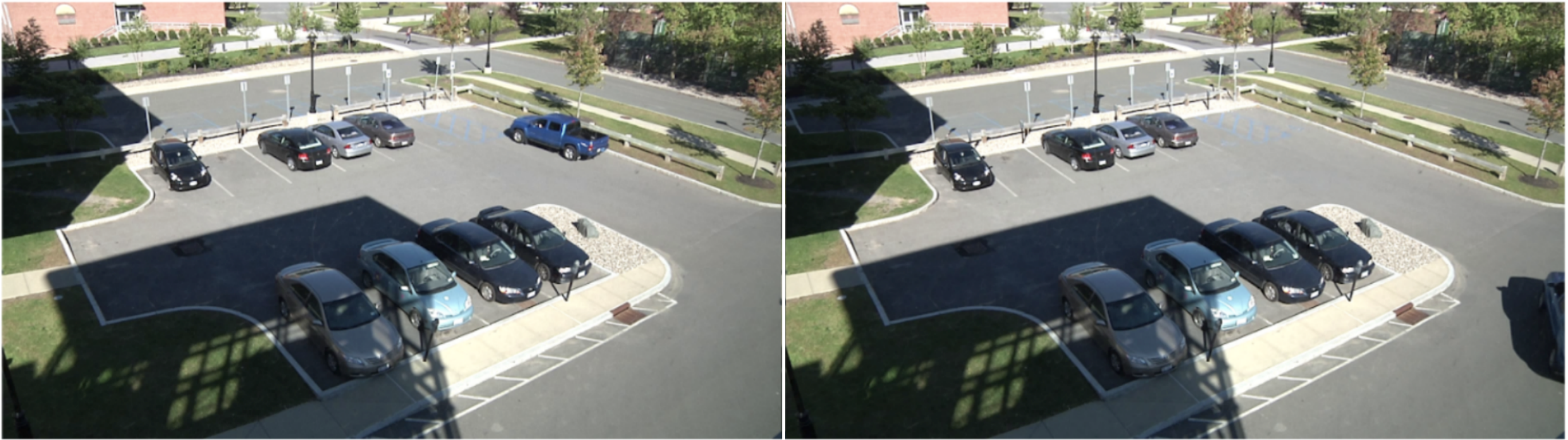} &
    The blue truck is no longer there. A car is approaching the parking lot from the right. \\

    <\textit{Tell me why the man points to the screen when talking to the child.}, {\color{xkcdOlive}\faImages\color{black}\ \textit{or} \color{xkcdRust}\faFilm\color{black}}> & \textit{Video Question Answering} \\
    \vspace{-0.275cm}
    \includegraphics[width=\linewidth,height=1.75cm]{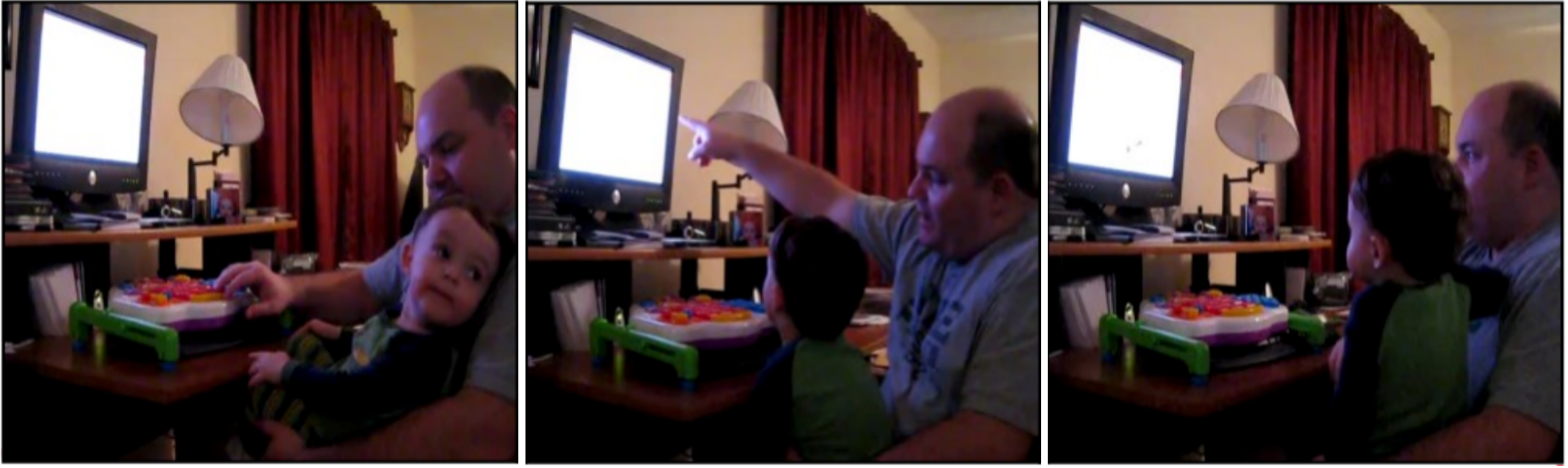} &
    To draw the child's attention. \\

    <\textit{Describe this video.}, \color{xkcdRust}\faFilm\color{black}> & \textit{Video Captioning} \\
    \vspace{-0.275cm}
    \includegraphics[width=\linewidth,height=1.75cm]{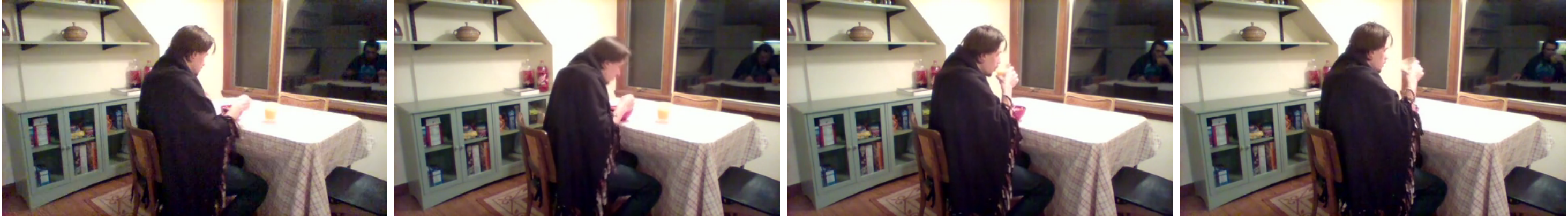} &
    The person is sitting at the dining room table wrapped in a blanket.  The person is eating cereal and drinking orange juice. \\

    <\textit{Generate a description for this movie clip that complements the dialogue.}, \color{xkcdRust}\faFilm\color{black}> & \textit{Movie Auto Audio Description} \\
    \vspace{-0.275cm}
    \includegraphics[width=\linewidth,height=1.75cm]{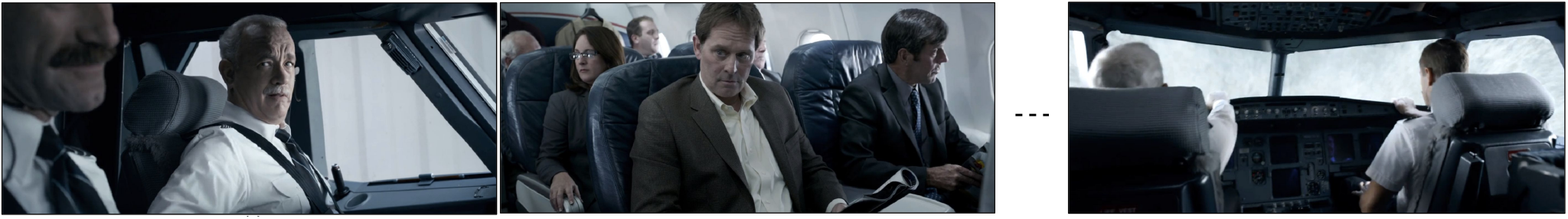} &
    Sully adjusts his seat harness. A male passenger looks up from his magazine ... Sully sticks out an arm as the jet bellies down onto the river.\\

    <\textit{Write a story for this image sequence.}, \newline{\color{xkcdOlive}\faImages\color{black}\ \textit{or} \color{xkcdRust}\faFilm\color{black}}> & \textit{Visual Storytelling} \\
    \vspace{-0.275cm}
    \includegraphics[width=\linewidth,height=1.75cm]{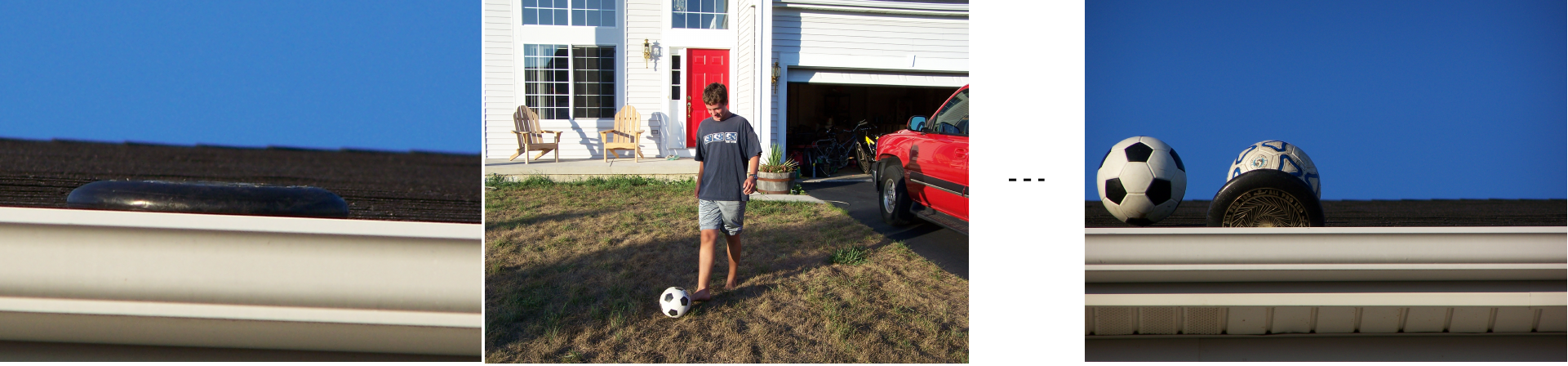} &
    A discus got stuck up on the roof. Why not try getting it down with a soccer ball? ... It didn't work so we tried ... are all stuck on the roof.
  \end{tblr}
  \Description{On the left, this figure shows a sequence of images or video frames positioned side-by-side. On the right, text corresponding to the visual inputs is provided.}
  \caption{Five tasks exemplifying the problem of text generation from visual events unfolding over time. For each task, we denote the visual input data type---sequences of images or video frames---using \color{xkcdOlive}\faImages\;\color{black}and \color{xkcdRust}\faFilm\;\color{black}respectively. In all cases, the output is natural language text.}
  \label{fig:review_tasks_examples_main}
\end{figure}

To ground our argument, we examine five seemingly distinct tasks, each of which we claim exemplifies this broader multimodal challenge. Our aim is to provide a survey of modeling approaches and evaluations proposed for a few selected, relevant tasks and to illustrate the general problem of generating language from visual events. The tasks we consider are \textit{Change Captioning}~\cite{cc_spot_the_diff}, \textit{Video Question Answering}~\cite[\textit{Video QA};][]{videoqa,miqa}, \textit{Video Captioning}~\cite{vc_task}, \textit{Movie Auto Audio Description}~\cite[\textit{Movie Auto AD};][]{maad1,maad2}, and \textit{Visual Storytelling}~\cite{vist}. Figure~\ref{fig:review_tasks_examples_main} presents an illustration and a prototypical example for each of the five tasks, drawn from publicly available datasets. By analyzing them through a shared lens, we highlight common modeling and evaluation challenges that are often overlooked when tasks are considered in isolation. Yet, we recognize that these tasks also differ in crucial ways that stem from their distinct communicative intents and objectives. For instance, crafting a coherent and imaginative narrative in \textit{Visual Storytelling} poses fundamentally different demands from pinpointing specific visual changes in \textit{Change Captioning}. These task-specific nuances, which can be ignored or flattened when focusing solely on shared technical challenges, underscore the need for specific modeling proposals and evaluation protocols, which is the second key argument of our paper.

Our goal is to promote a rethinking of how language generation from visual events is conceptualized, studied, and evaluated within multimodal NLP. In doing so, we aim to (i) articulate a unified framework for understanding multimodal tasks involving visual event sequences, (ii) identify key open questions and methodological gaps, and (iii) propose directions for future research that draw on insights from human cognition and linguistic theory, recognizing the communicative specificities of each task. We contend that improving models' ability to understand and generate language about visual events is timely and necessary, both for practical applications and for advancing our knowledge of how visual events are understood, interpreted, and narrated by humans and machines. The main contributions of this work are as follows:

\begin{itemize}[topsep=0pt]
  \item We offer a focused survey of representative modeling approaches, training paradigms, and evaluation methodologies used across five vision-to-language tasks involving visual events. By identifying shared design principles and methodological commonalities, we provide a unified lens that we hope will serve as a foundation for future research.
  \item We identify and articulate a set of core challenges that recur across tasks dealing with temporally ordered visual events. These include tracking and grounding of entities over time, modeling event continuity and causality, and maintaining coherence and cohesion in the generated language. We argue that these challenges are also the constructs of interest that evaluation methods should target.
  \item We show that, within the problem of generating language from visual events, different tasks vary along several features depending on their communicative intents. To this end, we provide a quantitative characterization of the tasks based on two dimensions---\textit{visual similarity} and \textit{textual consistency}---capturing the similarity between consecutive elements in the visual modality (images or video frames) and in the textual modality (sentences), respectively.
  \item We outline several promising research directions aimed at advancing models' ability to capture the complex temporal relationships between visual and linguistic modalities. These include more naturalistic and interactive task settings, modular architectures that combine task-specific components with general-purpose vision-language models (VLMs), and evaluation approaches that leverage interpretability methods.
\end{itemize}

In Section~\ref{sec:related_surveys}, we relate our work to existing surveys, which focus either on individual tasks or on general-purpose VLMs, and clarify its scope.

\section{Related Surveys and Scope}
\label{sec:related_surveys}

Several surveys have reviewed the individual tasks we consider in this paper, as well as the general-purpose VLMs that are increasingly used off-the-shelf to solve them. Image captioning, from which much of the modeling work in this domain originates \cite{vinyals2015show}, has been reviewed extensively \cite{stefanini2023show}. \textit{Video Captioning} (also referred to as video description) has been surveyed with a focus on methods, datasets, and evaluation metrics \cite{aafaq2019video}, while \textit{Video QA} has been reviewed in terms of its datasets, algorithms, and challenges \cite{zhong2022videoqa}. Similarly, existing reviews of \textit{Visual Storytelling} focus on generation techniques and related tasks \cite{oliveira2025vsg}, and \textit{Movie Auto AD} has been discussed in a recent review of generative technologies that are transferable to audio description \cite{gao2025ad}. As for \textit{Change Captioning}, existing reviews are limited to the remote-sensing domain \cite{zou2025rsicc}, and to the best of our knowledge, the task has not been surveyed in the general domain to date. At a broader level, surveys of general-purpose VLMs cover their architectures, pre-training objectives, and downstream tasks \cite{zhang2024vlmsurvey,yin2024mllmsurvey}, including video understanding \cite{tang2026vidllm}. Other surveys focus on benchmarks \cite{rd_eval2}, reasoning abilities \cite{mm_reasoning2}, and visually grounded language learning \cite{vg1}.

Our work differs from the above in three ways. First, existing surveys are organized either around a single task or around a specific class of models. In this paper, instead, we focus on the broader, more general problem of generating language from visual events unfolding over time, of which the tasks we consider are instances. Analyzing five seemingly distinct tasks through a shared lens allows us to identify methodological commonalities in modeling and evaluation (Sections~\ref{sec:models} and~\ref{sec:eval}), and to articulate a common set of challenges (Section~\ref{sec:challenges}) that are often overlooked when tasks are considered in isolation. Second, we complement the survey with a quantitative characterization of the tasks along two dimensions---\textit{visual similarity} and \textit{textual consistency} (Section~\ref{sec:data_properties})---which allows us to place the tasks at different positions in a shared space. Third, rather than providing an exhaustive account of the literature, we report a selection of recent approaches that illustrate both the main trends and the task-specific nuances stemming from the distinct communicative intents of the tasks. Readers interested in more extensive coverage of each task are referred to the surveys mentioned above and to the curated list of resources accompanying this paper.\footnote{A curated list of resources for the tasks discussed in this paper, including models, datasets, and evaluation metrics, is available at \url{https://github.com/akskuchi/nlg-from-visual-events}.}

Finally, we clarify the scope of our survey. Several active lines of research involving VLMs are related to, yet distinct from, the problem of language generation from visual events. Spatial intelligence, embodied robotics, and the use of VLMs in extreme scenarios (e.g., adverse weather or low-light conditions) primarily concern perception or action. In these settings, language, when involved, typically serves as an input or as a means of interaction rather than as the output to be generated. We therefore do not review these lines of research in this paper. We note, however, that the abilities discussed in Section~\ref{sec:challenges} are important prerequisites for making progress in these areas as well. For instance, an embodied agent that narrates or reasons about its surroundings needs to track entities over time and to ground underspecified language in its visual input. Moreover, the more naturalistic and interactive settings we advocate for in RD 1 (Section~\ref{sec:future_directions}) can serve as a bridge between language generation from visual events and embodied scenarios. We argue that extending our unified perspective to these settings is a promising avenue that should be explored in future work.

\section{Tasks and Datasets}
\label{sec:tasks_and_analysis}

In this section, we first define the five tasks we consider (Section~\ref{sec:task_definitions}) and provide an overview of the main datasets proposed for each of them (Section~\ref{sec:datasets}). We then quantitatively characterize the datasets used to train models for these tasks, in order to examine the extent to which they share similar underlying multimodal properties (Section~\ref{sec:data_properties}).

\subsection{Task Definitions}
\label{sec:task_definitions}

The five tasks we consider, illustrated in Figure~\ref{fig:review_tasks_examples_main}, are defined as follows. \textit{Change Captioning} requires generating a caption to describe semantic changes between a pair of related input images. This task is particularly relevant for applications involving surveillance or quality inspection of `before' and `after' visual data. The goal of \textit{Visual Storytelling} is to generate a fictitious, coherent narrative from a temporally-ordered sequence of images or frames. Applications of this task include digital storytelling and media creation. \textit{Video Captioning} concerns describing videos by summarizing them, while  \textit{Video QA} involves generating appropriate answers to questions about the entities and events in videos. Both tasks have a wide range of applications, such as content summarization on streaming platforms and interacting with virtual assistants. \textit{Movie Auto AD} involves generating narrative descriptions to complement the dialogues and soundtracks of movies. Enhancing the accessibility of movies and automating dubbing are some of the applications of this task.

\subsection{Datasets}
\label{sec:datasets}

\begin{table}[t]
    \centering
    \resizebox{\linewidth}{!}{
    \begin{tabularx}{\linewidth}{*{5}{>{\raggedright\arraybackslash}X}}
        \toprule
          \textbf{Dataset} & \textbf{Visual Input} & \textbf{Size} & \textbf{Domain} & \textbf{Output Text} \\
        \midrule
          \multicolumn{5}{c}{\textit{Change Captioning}} \\
        \midrule
          Spot-the-Diff \cite{cc_spot_the_diff} & Image pairs (video frames) & 13,192 pairs & Surveillance & Difference descriptions \\
          CLEVR-Change \cite{cc_duda} & Image pairs & 79,606 pairs; 493,735 captions & Synthetic 3D scenes & Generated change captions \\
          LEVIR-CC \cite{levircc} & Image pairs & 10,077 pairs; 50,385 sentences & Remote sensing & Change descriptions \\
        \midrule
          \multicolumn{5}{c}{\textit{Video QA}} \\
        \midrule
          MSVD-QA \cite{msvd_qa} & Videos (avg.\ 10\,s) & 1,970 videos; 50,505 QA pairs & Open domain & Generated one-word answers \\
          MSRVTT-QA \cite{msvd_qa} & Videos (avg.\ 15\,s) & 10,000 videos; 243,690 QA pairs & Open domain & Generated one-word answers \\
          NExT-QA \cite{nextqa} & Videos & 5,440 videos; 52,044 QA pairs & Daily activities & Multiple-choice and open-ended answers \\
        \midrule
          \multicolumn{5}{c}{\textit{Video Captioning}} \\
        \midrule
          MSR-VTT \cite{msrvtt} & Videos (41.2\,h in total) & 10,000 clips; 200,000 captions & Open domain & Single-sentence captions \\
          Charades \cite{vc_charades} & Videos (avg.\ 30\,s) & 9,848 videos; 27,847 descriptions & Indoor daily activities & Descriptions \\
          YouCook2 \cite{youcookii} & Videos (avg.\ 5.3\,min) & 2,000 videos & Cooking & Temporally localized step descriptions \\
        \midrule
          \multicolumn{5}{c}{\textit{Movie Auto Audio Description}} \\
        \midrule
          LSMDC \cite{lsmdc} & Movie clips & 202 movies; 118,114 sentences & Movies & ADs and script sentences \\
          MAD \cite{madv1} & Full-length movies (features only) & 650 movies; 384.6K sentences & Movies & ADs \\
          CMD-AD \cite{maad3} & Movie clips & 1,432 movies; 101,268 ADs & Movies & ADs \\
        \midrule
          \multicolumn{5}{c}{\textit{Visual Storytelling}} \\
        \midrule
          VIST \cite{vist} & Photo sequences & 20,211 sequences; 81,743 photos & Flickr albums & Stories and captions \\
          VWP \cite{vwp} & Sequences of 5--10 movie shots & $\sim$2K sequences; 12K stories & Movies & Character-grounded stories \\
        \bottomrule
    \end{tabularx}
    }
  \caption{A selection of datasets commonly used for the tasks considered in this paper, in chronological order within each task. For each dataset, we report the type of visual input, its size, the domain of the visual data, and the type of output text (AD: audio description).}
  \label{tab:datasets}
\end{table}

Table~\ref{tab:datasets} presents an overview of some of the main datasets proposed for each of the five tasks, including the ones we analyze in Section~\ref{sec:data_properties}. Rather than providing an exhaustive list, we select representative datasets that illustrate the diversity of the data both within and across tasks.

From this overview, we observe several trends. First, datasets have grown in scale and in the temporal span of their visual input, moving from short video clips (e.g., 10 to 30 seconds on average in MSVD-QA and Charades) to full-length movies in MAD \cite{madv1}. To obtain text at such scale, datasets often reuse existing text, such as audio descriptions aligned to movies \cite{madv1,maad3}, or generate it automatically, as for the change captions of CLEVR-Change \cite{cc_duda} and the question-answer pairs that MSVD-QA and MSRVTT-QA derive from video captions \cite{msvd_qa}. Second, more recent datasets tend to target specific abilities of models, e.g., causal and temporal reasoning in NExT-QA \cite{nextqa} or the grounding of characters in VWP \cite{vwp}. Third, the same visual sources are often used across tasks. For instance, MSVD-QA and MSRVTT-QA are derived from video captioning datasets \cite{msvd_qa}, and movies are used both for \textit{Movie Auto AD} and for \textit{Visual Storytelling} \cite{vwp}, which further supports our view of these tasks as instances of the same problem.

At the same time, current datasets have several limitations. Most of them contain English text only, and many are restricted to specific domains, such as surveillance, remote sensing, or movies, which may limit the generalization of models trained on them to other scenarios. Automatically generated text tends to be short and simple, e.g., all answers in MSVD-QA and MSRVTT-QA consist of a single word \cite{nextqa}, which limits its use for training and evaluating abilities such as textual coherence (Section~\ref{sec:challenges}). Moreover, copyright constraints can prevent the release of the raw visual data, as in the case of MAD, whose movies are only available as pre-extracted features \cite{madv1}. Beyond these characteristics, datasets for different tasks also differ in the properties of their visual and textual modalities, which we examine next.

\subsection{Properties of Datasets}
\label{sec:data_properties}

\begin{figure*}[t]
  \centering
  \subcaptionbox{Between consecutive images or video frames of 
  the visual modality.}
  {\includegraphics[width=0.49\textwidth]{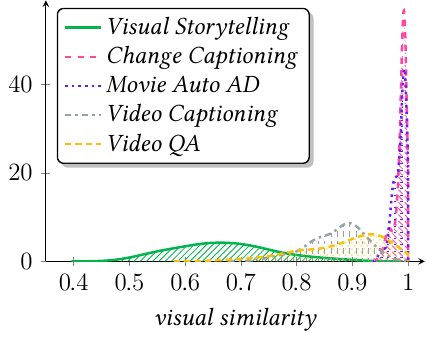}}
  \hfill
  \subcaptionbox{Between consecutive sentences in ground-truth text.}
  {\includegraphics[width=0.49\textwidth]{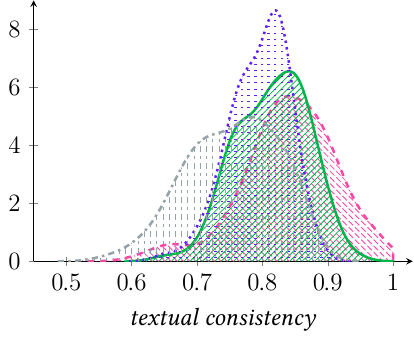}}
  \Description{}
  \caption{Similarity scores obtained for the tasks along the visual and textual dimensions. We excluded \textit{Video QA} from the \textit{textual} analysis due to the lack of multi-sentence datasets.}
  \label{fig:data_analysis}
\end{figure*}

Different tasks within this space may have different characteristics, due \textit{in primis} to their various communicative intents. Here, we briefly examine two possible features of datasets used to train models for the examined tasks: the similarity of the elements in the visual modality and the consistency of the textual modality.

The images or frames on the visual side can be more or less similar to each other, depending on the task. For instance, in the \textit{Change Captioning} task, where the goal is to localize and describe changes between a pair of images obtained from real-time surveillance cameras or large-scale remote-sensing snapshots, we expect the similarity between images within pairs to be generally high, because the two images depict the same scene and typically differ only by a few elements, such as a new car entering the scene. On the other hand, in tasks such as \textit{Visual Storytelling}, where the images (typically drawn from personal photo albums) depict events loosely forming an overarching narrative, we expect low similarity between consecutive images within each data sample (<visual sequence, text> pair). This can be due to the narrative-driven abrupt transitions between images in terms of characters or locations. Regarding the textual modality, we similarly expect that the consistency of consecutive sentences within each data sample will be higher or lower depending on the corresponding communicative intent of the task. By consistency, we do not mean logical consistency (i.e., the absence of contradictions between sentences), but rather the degree to which consecutive sentences are about the same content, such as the same entities, actions, or setting. For example, in the \textit{Movie Auto AD} task, where the language modality is intended to creatively complement the dialogues and soundtrack of the movie, we expect relatively low consistency between consecutive sentences within each data sample. Measuring this property allows us to distinguish tasks in which the text mostly describes the same content from tasks in which it develops over time, e.g., into a narrative. As we discuss in Section~\ref{sec:eval}, this distinction is important both for evaluation, since reference-based metrics are less appropriate for text with low consistency, and for the relative importance of abilities such as textual coherence.

To illustrate these properties, we quantitatively analyze the characteristics of five datasets corresponding to the sample tasks considered. Specifically, for each data sample, we operationalize \textit{visual similarity} by computing cosine similarity scores for each pair of consecutive images. We use the CLIP visual encoder \cite{clip}. Similarly, we operationalize \textit{textual consistency} by computing cosine similarity scores between CLIP text encoder embeddings of consecutive sentences in the corresponding ground-truth text of each data sample. For this study, we randomly select 100 instances per task from five datasets---VIST \cite{vist} for \textit{Visual Storytelling}, Spot-the-Diff \cite{cc_spot_the_diff} for \textit{Change Captioning}, Charades \cite{vc_charades} for \textit{Video Captioning}, MSVD-QA \cite{msvd_qa} for \textit{Video QA}, and MAD-v1 \cite{madv1} for \textit{Movie Auto AD}.

\begin{figure}[htbp]
  \centering
  \includegraphics[keepaspectratio,scale=1]{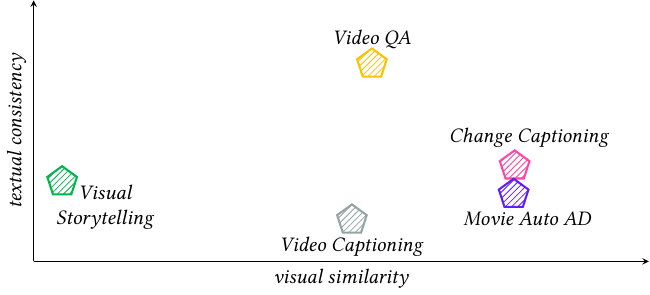}
  \Description{}
  \caption{Relative positioning of the tasks in the visual-textual shared space, based on the similarity scores obtained for the corresponding datasets. To illustrate, a task in the top right corner of the plot is both highly consistent at the textual level and highly similar at the visual level. Since no \textit{Video QA} dataset contains multi-sentence answers, we arbitrarily set its average \textit{textual consistency} value to $1$ to allow visualization.}
  \label{fig:shared_space}
\end{figure}

Figure~\ref{fig:data_analysis} shows the distributions of similarity scores obtained for each of the tasks along the visual and textual dimensions. In terms of \textit{visual similarity}, we observe that \textit{Change Captioning} and \textit{Visual Storytelling} obtain maximum and minimum scores, respectively, with other tasks ranging in between. In terms of \textit{textual consistency}, the differentiation is less evident. We observe that for the \textit{Movie Auto AD} and \textit{Video Captioning} tasks, consecutive sentences in the ground-truth text are relatively less consistent with each other. Using the average similarity scores across all data samples, we categorize the five tasks by placing them at different positions in the shared space between \textit{textual consistency} and \textit{visual similarity} (see Figure~\ref{fig:shared_space}).

This analysis shows that properties of data may differ significantly across tasks within the same problem space. While this analysis constitutes a small case study, we hope that it will encourage the community to explore the tasks further for identifying other features of the data that researchers aim to model. We believe that these and similar analyses of datasets have several benefits. First, they can be used to inform modeling and evaluation decisions within the shared space of NLG problems dealing with visual events. In Section~\ref{sec:eval}, we specifically discuss current approaches to evaluation. Furthermore, understanding the communicative intents behind specific tasks and analyzing properties of their data can modulate the importance of different abilities required for approaching them optimally. In Section~\ref{sec:challenges}, we identify key open challenges related to such abilities that are common across tasks and describe them in a unified manner.

Despite each task representing a unique point in this shared space---where the visual and textual data can be more or less similar---the core of our unified perspective is that any task involving language generation from visual events has to rely on a shared set of underlying model abilities. While the properties of datasets for these tasks diverge along the visual and textual dimensions, with the advent of transformers, the tasks are now fundamentally tackled in a similar manner using a unified model architecture. In the next section, we focus on the different components of this architecture and discuss how they are operationalized for different tasks.

\section{Model Architectures}
\label{sec:models}

Modeling approaches to text generation from visual events have evolved over time from being based on recurrent neural networks \cite[RNN;][]{lstm} to being transformer-based \cite{transformer}. More recent models directly leverage pre-trained large (vision)-language models (LLMs/VLMs), often in a zero-shot manner. In this section, we discuss this evolution and summarize the various state-of-the-art model architectures proposed for this problem. We ground our discussion in five representative tasks---defined in Section~\ref{sec:task_definitions}---that serve to exemplify the main trends.

These tasks originate from varied research communities that focus on different modalities (language \textit{vs.} vision) and place emphasis on different applications. However, as we argue in this section, it is possible to identify an underlying modeling approach common to tasks dealing with text generation from visual events. Architectures proposed for these tasks primarily comprise three modules---a vision encoder, a language decoder, and an intermediate module (typically referred to as the projector or adapter) for adapting visual information into contextualized representations for text generation. We describe these modules' functionality and review the common design principles across tasks. We also discuss how off-the-shelf pre-trained VLMs are currently being used to handle tasks involving text generation from visual events. We provide an overview of the models we reviewed for each task in Table~\ref{tab:models}.

\begin{table}[t]
    \centering
    \resizebox{\linewidth}{!}{
    \begin{tabularx}{\linewidth}{XXXX}
        \toprule
          \textbf{Model} & \textbf{Vision Encoder} & \textbf{Projector} & \textbf{Language Decoder} \\
        \midrule
          \multicolumn{4}{c}{\textit{Change Captioning}} \\
        \midrule
          CARD \cite{cc_card} & ResNet, Transformer$^\dag$ & \texttt{NA} & Transformer$^\dag$ \\
          SCORER+CBR \cite{cc_scorer} & ResNet, MH(S/X)A$^\dag$ & \texttt{NA} & Transformer$^\dag$ \\
          VARD-Trans \cite{cc_vard} & ResNet, Linear(s)$^\dag$ & \texttt{NA} & Transformer$^\dag$ \\
          DUDA \cite{cc_duda} & ResNet, RNN$^\dag$ & \texttt{NA} & RNN$^\dag$\\
        \midrule
          \multicolumn{4}{c}{\textit{Video QA}} \\
        \midrule
          ViLA \cite{videoqa_vila} & ViT, Transformer$^\dag$ & Q-Former$^\dag$ & {Flan-T5 XL} \\
          LLaMA-VQA \cite{videoqa_llamavqa} & CLIP-ViT-L & Linear$^\dag$ & \textsc{Llama} \\
          SeViLA \cite{videoqa_sevila} & ViT & Q-Former$^\dag$ & {Flan-T5 XL} \\
          FrozenBiLM \cite{videoqa_fbilm} & CLIP-ViT-L & Linear$^\dag$ & {DeBERTa-V2-XL} \\
        \midrule
          \multicolumn{4}{c}{\textit{Video Captioning}} \\
        \midrule
          Vid2Seq \cite{vc_vid2seq} &  {CLIP-ViT-L, Transformer$^\dag$} & \texttt{NA} & {T5-base} \\
          TextKG \cite{vc_textkg} & Transformer$^\dag$ & \texttt{NA} & Transformer$^\dag$ \\
          VTAR \cite{vc_vtar} & {InceptionResNetV2, C3D} & Transformer$^\dag$ & Transformer$^\dag$ \\
          ENC-DEC \cite{vc_task} & 3DCNN$^\dag$ & Attention$^\dag$ & RNN$^\dag$ \\
        \midrule
          \multicolumn{4}{c}{\textit{Movie Auto Audio Description}} \\
        \midrule
          MM-Narrator \cite{maad_mm_narrator} & CLIP-ViT-L & \texttt{NA} & {GPT-4} \\
          AutoAD-III \cite{maad3} & EVA-CLIP & Q-Former$^\dag$ & \textsc{Llama}~2 \\
          AutoAD-II \cite{maad2} & {CLIP-ViT-L} & \texttt{NA} & {GPT-2, MHXA$^\dag$} \\
          AutoAD \cite{maad1} & CLIP-ViT-L & {Transformer$^\dag$} & {GPT-2} \\
        \midrule
          \multicolumn{4}{c}{\textit{Visual Storytelling}} \\
        \midrule
          MCSM+BART \cite{vist_kg1} & ResNet, RNN$^\dag$ & \texttt{NA} & {BART} \\
          TAPM \cite{vist_tapm} & ResNet, FRCNN & \texttt{NA} & {GPT-2} \\
          KG Story \cite{vist_kg2} & FRCNN & \texttt{NA} & Transformer$^\dag$ \\
          GLAC Net \cite{vist_glacnet} & ResNet, RNN$^\dag$ & \texttt{NA} & RNN$^\dag$ \\
        \bottomrule
    \end{tabularx}
    }
  \caption{A selection of recent models proposed for the tasks involving visual events considered in this paper. We only report models that are end-to-end and task-specific, i.e., trained or fine-tuned for the task at hand. For each model, we report the underlying vision encoder and language decoder, as well as the projector module, when applicable (\texttt{NA}: Not Applicable). Components with $^\dag$ have been trained from scratch using only the datasets available for the corresponding task.}
  \label{tab:models}
\end{table}

\subsection{Vision Encoder}

The primary purpose of a vision encoder in vision-to-language tasks is to extract information from the visual input and to optimally encode it into a contextual representation that guides language generation. To achieve this, encoders in the proposed models follow multiple steps, some of which are common across language generation tasks from visual events. First, a pre-trained vision model is utilized for extracting feature representations of the raw input sequences of images or video frames. Earlier approaches used vision models based on convolutional neural networks (CNNs) such as ResNet \cite{resnet} or R3D \cite{resnet3d} that are primarily pre-trained for object or action recognition using large amounts of image or video data. Most of the recent models across tasks use transformer-based vision models pre-trained for various image-only and image-text alignment objectives, e.g., CLIP-ViT-L \cite{clip}. We note that besides the primary input sequence of images or video frames, models proposed for some of the tasks, e.g., \textit{Movie Auto AD}, utilize additional input data such as close-ups of characters in the movie clips \cite[\textit{exemplars};][]{maad2}. Furthermore, the TAPM \cite{vist_tapm} model proposed for the \textit{Visual Storytelling} task utilizes Faster R-CNN \cite{faster-rcnn} for extracting such local entity-level features alongside the global image-level features from ResNet.

Following the extraction of visual features using pre-trained vision models, most vision encoders comprise an internal sequence-encoder component for learning relationships and dependencies between the individual image or frame-level features at different temporal positions. Some models implement this step either using RNNs or a transformer network with multi-head self-attention for learning temporal relationships and a position-encoding mechanism for tracking the order of entities or events in the visual sequence.

\begin{figure}[h]
  \centering
  \includegraphics[keepaspectratio,scale=1.1,width=\textwidth]{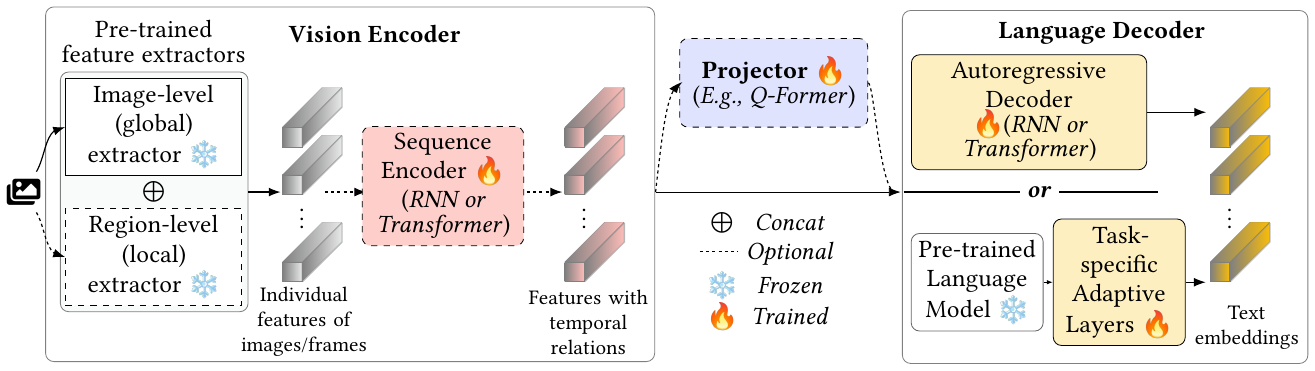}
  \Description{}
  \caption{Outline of the architecture common across modeling approaches for tasks involving language generation from visual events.}
  \label{fig:model_arch}
\end{figure}

Beyond these common components across tasks, vision encoders may also contain additional task-specific steps for capturing the visual information in a way that suits the task's objective better. For instance, the ViLA model \cite{videoqa_vila} for the \textit{Video QA} task utilizes a learnable Frame-Sampler component to efficiently select a small subset of frames that are most likely to contain the relevant information needed to answer the question. Another example with a task-specific step is the MCSM+BART model \cite{vist_kg1} for \textit{Visual Storytelling}, in which the initial set of image objects or `concepts' is expanded using an external knowledge graph for generating diverse and informative stories. Despite these task-specific steps, we find that the vision encoder module in the architectures proposed for the various  language generation tasks dealing with visual events shares a common set of components that are broadly outlined in Figure~\ref{fig:model_arch}.

\subsection{The Vision-to-Language Bridge}

Some vision-to-language model architectures use an intermediate module that bridges the input and output modalities for effectively conditioning text generation on the extracted visual features. Different models operationalize this module with different degrees of complexity. Earlier approaches condition the text generation process by directly fusing vision encoder outputs with the language decoder input \cite{vist_glacnet}. Some architectures employ cross-attention mechanisms to focus on the relevant parts of the visual features at various temporal positions during decoding \cite{vc_task}. However, approaches that adopt pre-trained models---e.g., CLIP-ViT-L \cite{clip} as the visual model---tend to employ learnable intermediate layers for aligning and converting outputs of the vision encoder into a format that the language decoder can understand.

In some of the proposed models, this intermediate module is a single linear layer that transforms the visual features into a common shared space, which can then be used by the language decoder \cite{videoqa_llamavqa,llava}. In other models, advanced transformer-based projectors such as a Q-Former \cite{blip2} are used based on their ability to leverage cross-modal interactions effectively \cite{maad3}. In essence, Q-Former uses learnable query vectors that are pre-trained to attend to both visual and textual representations, enhancing its ability to generalize and perform well (relative to a single linear layer) across different tasks. Besides these popular methods for adapting multimodal information, some approaches make use of graph neural networks for capturing relationships between objects in the images at different temporal positions and words in the corresponding sentences of the text \cite{vc_gnn}.

While there is no definitive way to design this intermediate module, recent work has compared the two approaches, i.e., using cross-attention between modalities or using a multimodal projector for transforming vision encoder features into the language space, and found that the latter performs better when the pre-trained backbones are also trained, whereas cross-attention is better when they are frozen \cite{what_matters}. This suggests that using intermediate multimodal projectors can help in extracting meaningful visual representations.

\subsection{Language Decoder}

After encoding and adapting the visual information, models employ a language decoder component for text generation. The decoder can either be learned from scratch or consist of a pre-trained language model with additional trainable task-specific layers. Figure~\ref{fig:model_arch} summarizes the different ways in which this step is operationalized across tasks in the proposed architectures. Earlier models learn an RNN by initializing it with the visual context embedding from the previous steps \cite{vc_task,vist_glacnet}. The decoder then typically follows a `teacher forcing' strategy during training to generate one word at a time autoregressively.

Subsequent models have replaced RNNs with the transformer architecture owing to its computational scalability and efficiency in handling long context windows. Besides the initial word embedding layer and the position encoding step (for maintaining information about the input sequence token order), a transformer decoder is typically made up of multiple identical blocks. Each block comprises a multi-head self-attention layer for modeling intra-sentence relations (between the words) and a multi-head cross-attention layer for learning relationships between representations of each word and the outputs of the visual encoder or projector. For instance, in the \textit{Change Captioning} task, this refers to conditioning each word in the caption on vision encoder outputs (denoted as `difference-representations').

Instead of training the decoder from scratch, some approaches use language models such as GPT-2 \cite{gpt2} and \textsc{Llama}~2 \cite{llama2}, which are pre-trained on large text-only corpora. The pre-trained language models are either used directly for generation by freezing their parameters \cite{maad1,maad2,maad3}, or adapted by inserting and fine-tuning additional adaptive layers on top of them for ensuring relevance of the generated text to the downstream task of interest \cite{vist_tapm}. We also note that some models incorporate information from external knowledge bases into the decoder module to improve coherence and factuality of the generated text, e.g., TextKG \cite{vc_textkg} for the \textit{Video Captioning} task and KG Story \cite{vist_kg2} for the \textit{Visual Storytelling} task. This approach has been shown to improve models in terms of correctness of answers, relevance of descriptions, and expressiveness of generated narratives. On the other hand, methods relying on external knowledge face several practical limitations, such as retrieval latency and model scalability. Furthermore, for applications in some domains (e.g., \textit{medical}, \textit{financial}), interacting with external sources may not be feasible due to privacy or security constraints.

\subsection{Off-the-shelf Pre-trained VLMs}
\label{sec:4_4}

The standard model architecture we have discussed so far is also generally present in powerful general-purpose foundation VLMs, which are pre-trained on several tasks using large amounts of data. Being multi-purpose by design, these models can be used directly for language generation tasks dealing with visual events. Their pre-training process typically happens in two stages---self-supervised alignment training and visual instruction tuning. During the first stage, only the parameters of the intermediate module connecting both unimodal backbones are updated (commonly using paired image-text data), using a next-token prediction (captioning) objective. In the second stage, models are instruction-tuned using multi-turn conversations obtained for visual data either through crowd-sourcing or by leveraging models such as GPT-4 \cite{gpt4}.

Contrary to task-specific modeling approaches, these pre-trained VLMs are simply prompted (typically in a zero-shot manner) using visual tokens accompanied by task-specific instructions. Recent work has tested the ability of some of these pre-trained models to generate language from visual events \cite{video-llava, mplugowl3, maad4, nytws}. These multi-purpose models have been shown to be on par with, or in some cases even outperform, models that are designed and trained specifically for those tasks. Table~\ref{tab:VLMs} outlines the key components of some pre-trained VLMs used in a zero-shot manner for different vision-to-language tasks dealing with visual events.

\begin{table}[ht]
    \centering
    \resizebox{\linewidth}{!}{
    \begin{tabularx}{\linewidth}{XXXX}
        \toprule
          \textbf{Model} & \textbf{Vision Encoder} & \textbf{Projector} & \textbf{Language Decoder} \\
        \midrule
          Qwen2.5-VL \cite{qwenvl} & ViT & Linear & Qwen2.5 \\
          DeepSeek-VL \cite{deepseekvl} & SigLIP, SAM-B & Linear (\#2) & DeepSeek \\
          LLaVA-NeXT \cite{llava} & CLIP-ViT-L & Linear & Mistral \\
          Video-LLaMA \cite{video_llama} & ViT-G/14 & Q-Former & \textsc{Llama} \\
          mPLUG-Owl3 \cite{mplugowl3} &  SigLIP SO (400M) & Linear & Qwen2 \\
          Video-LLaVA \cite{video-llava}& OpenCLIP ViT-L & Linear (\#2) & Vicuna v1.5\\ \bottomrule
    \end{tabularx}
    }
  \caption{A selection of pre-trained multi-purpose VLMs that are used off-the-shelf for various NLG tasks involving visual events.}
  \label{tab:VLMs}
\end{table}

Despite progress in terms of architectures and learning procedures, our ability to rigorously assess models remains severely limited, which has resulted in a critical evaluation bottleneck. This is because most of the existing evaluation approaches rely primarily on text overlap and tend to ignore the aspect of cross-modal alignment. As a result, most metrics fail to measure whether a model can ground expressions in the visual information, track entities over time, or reason about intentions of characters.

\section{Evaluation Constructs and Challenges}
\label{sec:eval}

Automatically generated text is notoriously difficult to evaluate. Most of the methods currently used to evaluate text describing visual events are borrowed from simpler NLG scenarios, and hence are not sufficient to fully understand the quality of language generated from visual input unfolding over time. These methods range from using traditional \textit{n}-gram matching metrics to obtaining human judgments and ratings to, more recently, using off-the-shelf pre-trained VLMs assessing the generated output. In this section, we first broadly classify evaluation methods into two main categories---automatic and human evaluation---and discuss several metrics widely used for vision-to-text tasks, along with the rationales for using them and the constructs they target, i.e., the qualities of the generated text or the abilities of models that they are intended to measure. Through this discussion, we also underline how the current approaches are lacking when it comes to assessing models with respect to the data features and the particular challenges relevant to the problem of language generation from visual events. We then describe these key challenges and argue that they are the constructs of interest that evaluation methods should target across all five tasks (Section~\ref{sec:challenges}).

\subsection{Automatic Evaluation}
\label{sec:5_1}

To computationally assess the quality of model-generated text along different aspects, several automatic metrics have been proposed. While some metrics rely on descriptions, narrations, or answers provided by human annotators, others are reference-free and assess model outputs independently of the ground-truth data. The former reference-based metrics typically neglect the visual modality, while reference-free metrics are more likely to address this limitation by proposing approaches that evaluate a model output in relation to the visual input.

\paragraph{Reference-based metrics}
Very often, tasks involving language generation from visual events assess a model-generated candidate text by comparing it to corresponding human-written references. Specifically, traditional metrics that were originally designed for evaluating machine translation and text summarization tasks---BLEU \cite{bleu}, METEOR \cite{meteor}, and ROUGE \cite{rouge}---are used to measure precision and recall of overlapping $\mathit{n}$-grams between the candidate and a reference text. Usually, metrics such as CIDEr \cite{cider} and SPICE \cite{spice}, which have been specifically developed for the evaluation of image and video captioning, are also used in conjunction with the above three metrics.

All the above-mentioned metrics rely on direct raw text comparisons of ground-truth references and model outputs. To go beyond exact \textit{n}-gram matches, embedding-level reference-based evaluation metrics such as WMD \cite{wmd}, BERTScore \cite{bertscore}, and ViLBERTScore \cite{vilbertscore} have been proposed. Recent work for \textit{Video Captioning}, \textit{Video QA}, and \textit{Movie Auto AD} has used these metrics to measure the similarity between candidate and reference embeddings, obtained by projecting corresponding text into a common pre-trained semantic space \cite{vc_swinbert,maad2,maad3}. In essence, reference-based metrics target the similarity between candidate and reference texts, either in terms of overlapping \textit{n}-grams or in terms of meaning, which only indirectly reflects the quality of the generated text.

\paragraph{Reference-free metrics}
Comparing model-generated text to ground-truth references, typically provided by crowd-workers or scraped from the internet, has various limitations when applied to the context of visually grounded language generation. First, most reference-based metrics do not account for the visual modality upon which the generated text is conditioned. Second, reference-based metrics are generally inappropriate to accurately evaluate language that is not merely descriptive, but may involve creative or narrative concepts. This is the case for various tasks involving visual events, such as \textit{Video Captioning}, \textit{Movie Auto AD}, and \textit{Visual Storytelling}, where the data shows a low degree of textual consistency between consecutive sentences, as shown in Figure~\ref{fig:shared_space}.

On the language-only level, various reference-free metrics---not relying on any reference text---such as MAUVE \cite{mauve} and UNION \cite{union} have been proposed. However, for visually conditioned text generation tasks, the shift from reference-based to reference-free evaluation is still relatively recent, with a persistent emphasis on reference-based $\mathit{n}$-gram metrics to date. That said, various reference-free metrics have been recently proposed to assess different aspects of evaluation that are important for generation tasks that specifically concern visual events. For instance, metrics such as CLIPScore \cite{clipscore} and GROOViST \cite{groovist} have been used for evaluating visual grounding---the degree of alignment between the generated text and the visual input---in \textit{Video Captioning} and \textit{Visual Storytelling} tasks. Similarly, the RoViST \cite{rovist} suite of metrics has been proposed to assess coherence, the extent of repetition, and visual grounding of the generated text. Yet other metrics, such as CRITIC \cite{maad3} and `character matching' \cite{vist_cm}, are designed to evaluate task-specific aspects such as the accuracy of references to characters in the model outputs. These metrics are among the few that explicitly target abilities specific to language generation from visual events, such as visual grounding, textual coherence, and, to some extent, entity tracking (see Section~\ref{sec:challenges}).

Finally, there is currently an increasing adoption of, and reliance on, using pre-trained LLMs and VLMs as judges \cite{llmeval2}. Essentially, these pre-trained general-purpose models are prompted to score or rate a model-generated response along any of the evaluation dimensions of interest, such as fluency or relevance (e.g., \textit{`How fluent is the generated text on a scale of 1 to 5?'}). However, the effectiveness and reliability of this approach are still debated \cite{llmeval1}.

\subsection{Human Evaluation}
Given the current state of automatic evaluation, some tasks, such as \textit{Video Captioning} and \textit{Visual Storytelling}, rely on human evaluation to accurately determine the quality of the model-generated text. This process involves recruiting online crowd-workers who are native or proficient speakers of the target language. Depending on the type of data and communicative specificity of the task, annotators with expertise and familiarity with terminology relevant to the corresponding domain (e.g., medical or sports videos) might be preferred.

Participants of these evaluation efforts are provided with a set of task-specific rubrics along with representative examples required for judging the model outputs. They are asked to assess the overall quality of model-generated outputs either independently (per sample) \cite{groovist} or relative to outputs from other models \cite{rovist}. Alternatively, evaluators might be required to provide ratings for various criteria ranging from broad (e.g., text conciseness, fluency, grammatical correctness) to specific (e.g., factuality, hallucinations, expressiveness). The obtained scores are usually compared pairwise to rank models appropriately.

Some pre-trained VLM frameworks, such as LLaVA-RLHF \cite{llava_rlhf}, leverage this qualitative feedback to optimize model parameters for learning to generate human-preferred text. Although human evaluation is still indispensable for several tasks, it is also expensive, time-consuming, and challenging. Because of this constraint, the community remains overly reliant on inadequate automatic metrics. Defining clear evaluation protocols for ensuring the reliability and quality of human judgments is an active research area \cite{human_eval_protocols1,human_eval_protocols2}.

Overall, the evaluation methods reviewed above primarily target general qualities of text, such as fluency, relevance, or similarity to references, and only a few of them target abilities that are specific to language generation from visual events. For both automatic and human evaluation, we argue that it is essential to first identify the key abilities relevant to the task at hand before developing metrics or questionnaires to comprehensively test models. We turn to such key challenges in the next subsection, where we treat each of them as a construct of interest and discuss the limitations of current evaluation methods in assessing it.

\subsection{Key Challenges as Evaluation Constructs}
\label{sec:challenges}

As mentioned in the Introduction, tasks involving language generation from visual events present a unique set of challenges. These involve modeling the intricate relationship between temporally ordered visual events and the structure, content, and function of the language used to interpret, describe, or narrate them. We argue that solving these tasks requires models to be capable of identifying and managing such intricacies, leveraging the temporal, causal, and discourse-level patterns inherent in both visual data and language, as well as the many ways in which these aspects interact. In this subsection, we identify and describe five main open challenges and discuss how they are key to various tasks. We treat each challenge as a construct of interest---an ability that models need and that evaluation methods should therefore target---and outline the limitations of current evaluation methods in assessing it. Since these constructs are shared across all five tasks, they support our view that the tasks are specific instances of the same general problem. This unified perspective is a novel contribution of our work, which we believe can help the community to find common strategies for making progress across tasks.

\paragraph{Entity tracking}
Identifying and tracking entities is an important requisite for accurately interpreting actions and relationships between them. This has been underlined by several works in both language understanding \cite{et_lm1,et_lm2} and computer vision domains \cite{et_vm1,et_vm2}. In our multimodal tasks, the availability of input signals from two modalities makes the aspect of disambiguating entities more challenging. For instance, in the \textit{Movie Auto AD} or \textit{Visual Storytelling} tasks where the visual input is heterogeneous, entities in the visual input tend to `disappear' in some of the images or video frames at the intermediate temporal positions, while still being actively referenced in the textual input at the corresponding positions. \citet{groovist} have denoted such cases as being \textit{temporally misaligned}. Traditional text-comparison-based approaches cannot evaluate how well a model resolves such temporal misalignments. To track entities accurately under such temporal misalignment, it is necessary not only to learn causal relations between people and objects in each of the modalities at corresponding positions, but also to obtain a cross-modal, cross-temporal representation of all the relationships relevant to the scene and the overarching narrative. For the \textit{Change Captioning}, \textit{Video Captioning}, or \textit{Video QA} tasks, in which the input images or video frames are typically similar to each other, it is crucial to differentiate meaningful semantic entities and their changes from various distractions. While in \textit{Change Captioning}, viewpoint or illumination changes are considered distractors and discarded, in the \textit{Video QA} task, entities relevant to answering the question need to be differentiated from others for accurate tracking. Effective tracking of entities would therefore require accounting for changes in appearance (including disappearance), capturing interactions, and correctly identifying occlusions.

\paragraph{Visual grounding}
Humans acquire language through perception and interaction with the environment \cite{vg0,lang_acquisition}, which enables them to seamlessly ground language in visual data. Over the years, a great deal of work has been proposed to adapt the architecture and learning process of vision-language models for acquiring visual grounding \cite{vg1}. However, there are still significant challenges for achieving human-level grounding using computational models, and this becomes more apparent when looked at from the perspective of the tasks under investigation here. For instance, in \textit{Visual Storytelling} or \textit{Movie Auto AD}, the language used is often creative or abstract, including various sources of underspecification, such as context-dependent adverbs like `\textit{there}',  `\textit{often}', or `\textit{today}' (`a photographer over \textit{there} clicks\dots'). Recently, vision-language models have been shown to struggle with underspecified language, due to the challenges related to \textit{grounding} underspecified terms to regions in the visual content \cite{semantic_underspecification}. Moreover, the amount of language informativeness---i.e., the degree of information required for identifying the correct object \cite{vg2}---could be inadequate in tasks such as \textit{Video QA}, particularly with the presence of confounding entities in various frames (e.g., input video of a football match and a question: \textit{`What is the color of the card the referee is holding?'}). In addition, when grounding objects, models have often struggled to reliably capture spatial relationships \cite{vg3}. Therefore, a proper evaluation of grounding requires accounting for spatial/embodied intelligence, as human grounding extends to physical spaces. To summarize, the language used in these tasks might not be merely descriptive but rather complementary to the content depicted in the corresponding images or videos. This makes visual grounding challenging without access to relevant additional external knowledge. Currently, most reference-based metrics cannot assess this construct, as they ignore the visual modality.

\paragraph{Knowledge integration}
To meet the communicative specificity of each task dealing with visual events, models would need to use additional information beyond what is available in the visual input. When performing \textit{Video Captioning} or \textit{Video QA} on specific domains such as \textit{news}, the input video might not contain all the aspects needed to correctly describe its contents or answer questions about it \cite{vc_kg_task,videoqa_kg_task}. To address this, various approaches often rely on using large pre-trained general-purpose models or external knowledge bases such as ConceptNet \cite{concept_net} to retrieve both factual and commonsense information. This method is commonly referred to as retrieval-augmented generation \cite[RAG;][]{rag}. Besides being sources for missing information, external knowledge bases are also leveraged for enriching the generated text with social or cultural contexts \cite{knowledge_integration1,knowledge_integration2}. For instance, in \textit{Visual Storytelling}, some approaches use recognized visual objects in the input to retrieve concepts from external knowledge graphs for generating more engaging and figurative stories (e.g., the concept of `\textit{graduation ceremony}' following the detection of an `\textit{academic gown}' object).

The process of integrating external knowledge has various challenges, ranging from selecting and retrieving relevant knowledge to accurately representing and using it during text generation. Robust retrieval systems that can holistically extract the essence of image sequences or video frames, including the various entities and their interrelationships, are required. Typically, the retrieved knowledge is concatenated with input representations, which are then used for generating text either through fine-tuning \cite{kg_finetuning} or by prompting general-purpose VLMs \cite{kg_promptbased}. However, this approach might lead to models either over- or under-exploiting the retrieved knowledge, potentially leading to incoherent text \cite{rag_issues}. To address this, fusion mechanisms that can effectively balance information from representations of both input data and the retrieved knowledge have yet to be developed. Furthermore, retrieving relevant knowledge from increasingly large databases could be computationally expensive, especially in scenarios where visual events unfold over time. Finding methods to optimize retrieval components to improve efficiency is an active research area. Integrating external knowledge also poses a challenge in that evaluation methods can struggle to distinguish between hallucinations and valid external knowledge.

\paragraph{Textual coherence}
Coherence is the property of text that refers to the ordering of its constituents (words, sentences) and how they relate to each other \cite{coh1}. Coherent text should have a consistent logical structure in which the events, interactions, and relationships between various elements are ordered in a meaningful way. It is an important aspect of discourse and has been studied extensively in neural language generation \cite{coh2}. For several NLG tasks involving visual events, it might be challenging to ensure that multiple sentences in the generated text are locally coherent. In \textit{Movie Auto AD} and \textit{Visual Storytelling}, where there are multiple characters and various interactions unfolding across the sequence of images or video frames, it is often difficult for models to balance between selecting the visual information and representing it cohesively using language \cite{coh3}. This challenge is more apparent in the \textit{Visual Storytelling} task, in which models are expected to keep track of multiple things, such as emotional arcs or motivations of the characters and the overarching narrative \cite{nytws}. In such tasks, there is a tension between coherence and creativity (cf.\ low \textit{textual consistency} of \textit{Visual Storytelling} data in Figure~\ref{fig:shared_space}). Assessing this requires metrics that account for more than \textit{n}-gram overlaps to evaluate long-term narrative structure. There is increasing work in unimodal text-only storytelling suggesting how using concepts of narratology \cite{narrativity1,narrativity2}, such as Genette's triangle \cite{narrativity_genette}, can potentially aid models in generating stories with engaging and coherent structures. However, how these theories can be applied to multimodal scenarios where the generated text needs to be consistent with visual input is still overlooked. We argue that this is a promising avenue that should be explored in future work.

\paragraph{Theory of mind}
Theory of Mind (ToM), which is considered the basis of human social cognition \cite{tom1}, is described as the ability to understand and make inferences about the mental states (e.g., beliefs, intentions, and desires) of other people or living beings. In the context of the tasks explored in this paper, ToM refers to the ability of models to go beyond merely recognizing actions and to reason about the mental states of entities depicted in the visual input. A recent benchmark for the \textit{Video QA} task focuses on ToM and shows that existing models lack this skill \cite{videoqa_bdiqa}. We argue that ToM is relevant for all the tasks involving language generation from visual events and that it is closely connected to the other challenges and abilities discussed so far. For instance, in tasks such as \textit{Visual Storytelling}, to causally connect heterogeneous images in the input sequence, models need to be equipped with different reasoning abilities pertaining to emotions, social perceptions, and intentions. This enables the generation of stories that reflect actions and mental states of the characters beyond the literal interpretation of the visual data. This is different in \textit{Movie Auto AD}, where attributing mental states to characters carries the risk of providing subjective descriptions. Professional audio description guidelines, for instance, recommend describing facial expressions, body language, and reactions, especially when these contradict the dialogue, while avoiding opinionated descriptions \cite{ad_guidelines}. In this task, ToM should therefore inform what is relevant to describe (e.g., a reaction that contradicts what a character says), rather than be expressed as explicit attributions of mental states.

None of the automatic metrics discussed in Section~\ref{sec:5_1} targets this construct, which is currently assessed only through benchmarks. Recently, several ToM benchmarks have been proposed to assess general-purpose VLMs \cite{tom2,tom3} along different aspects, such as temporal localization of emotions, intentionality-understanding, and perspective-taking. These studies found that only models that are fine-tuned on curated ToM datasets exhibit any reasoning abilities, albeit not aligning with the well-established ToM frameworks explaining human social cognition. Such curated data is scarcely available, and it is currently an open question what alternative architectures or training objectives would enable models to obtain the required ToM abilities. Once again, we claim that future work in this domain should pay more attention to this topic, as mastering ToM will likely lead to models that can better solve tasks involving visual events unfolding over time.

\section{Research Directions}
\label{sec:future_directions}

As discussed in previous sections, the problem of generating text from a sequence of temporally ordered images or frames is a challenging one, and relevant to several downstream tasks and applications. Here, we outline several research directions (RDs) that, we argue, are key to advancing model development and evaluation in this domain. Whereas the key challenges discussed in Section~\ref{sec:challenges} are abilities that models need to acquire, the directions we propose here concern changes to how tasks are defined (RD 1), how models are designed (RD 2), and how they are evaluated (RD 3).

\paragraph{RD 1: Towards more naturalistic scenarios}
Many of the tasks we consider in this work have real-world applications. For instance, solutions to the \textit{Change Captioning} task can be used for assisted surveillance and for tracking changes in digital media assets \cite{cc_spot_the_diff}. In the \textit{Movie Auto AD} task, models are required to generate descriptions that complement information in the original audio dialogue and soundtrack, for improved accessibility to visually impaired users and for enhancing the visual experience of sighted users \cite{maad2}.

However, many day-to-day human-centered scenarios involve personalizing to various contexts or situations. We argue that existing tasks in this domain, in their definitions and settings, do not fully reflect this aspect. Tailoring model-generated descriptions or narrations to the perspective of end-users requires task settings in which models and humans can interact iteratively. Such settings would enable the incorporation of human expectations and communicative contexts, which typically tend to be dynamic in real-world applications. To this end, we advocate for variations of existing tasks where models can learn to contextualize and reason through interactions with other agents (humans or other models) for generating stories, descriptions, or answers. In vision-to-language tasks involving single images, various works have explored controlled settings in which models are expected to generate text adhering to a specific style \cite{rd1_2,rd1_0} or user preferences \cite{rd1_3,rd1_1}. Models developed for solving these variations of tasks can be directly deployed to meet the requirements of real-world use cases. However, in tasks that require generating text from multiple images or video frames, there has been limited work concerning the abilities of models to control for the style of the generated output \cite{rd_task1}. We advocate for an exploration of such settings in language generation tasks dealing with visual events, for enabling the development of models that can adapt and cater to various points of view or genres dynamically.

\paragraph{RD 2: Are general-purpose VLMs all we need?}
As discussed in Section~\ref{sec:4_4}, VLMs that are trained on various general-purpose datasets are increasingly being used in these scenarios through prompting. Powerful open-source models such as Molmo \cite{molmo} and models optimized for multi-image scenarios such as Mantis \cite{mantis} are becoming increasingly available, suggesting that the trend of adopting them off-the-shelf for solving many vision-to-language tasks is widespread. General-purpose VLMs learn abundant information through multitask pre-training and have a modular design, making them suitable for many downstream tasks. Their modularity also enables seamless adaptation of VLMs to various novel domains (e.g., medical science) by updating only a small fraction of their parameters \cite{rd_model1}.

Despite the promising generalization of VLMs to certain tasks and domains, they have also been shown to be sensitive to prompts \cite{rd_model2_a,rd_model2_b} and biased towards the textual modality \cite{rd_model3}. To address these problems, recent work proposes various \textit{prompt engineering} techniques to facilitate inference-time adaptation of prompts to make them more suitable for the specific task of interest \cite{rd_model4,rd_model5}. On the other hand, task-specific model architectures consist of components designed to effectively address specialized aspects of the tasks, e.g., computing a \textit{difference representation} of the input image pair in \textit{Change Captioning}. We advocate for modular modeling approaches that bring together efficient task-specific components and combine them with the powerful foundational VLMs. Furthermore, we argue that using graph-based architectures and memory-based modules akin to \citet{rd_modular1,rd_modular2} would result in improved tracking of entity positions and relationships, and enable models to assign saliency to memorable events in tasks like \textit{Visual Storytelling} or \textit{Movie Auto AD}.

\paragraph{RD 3: Improving and rethinking evaluation}
In Section~\ref{sec:eval}, we discussed various approaches for evaluating model outputs in this domain. While human evaluation is impractical for conducting large-scale assessments, existing automatic evaluation metrics are limited in terms of fully capturing the abilities of models. Increasingly, various benchmarking datasets are being proposed to assess models along different axes important for grounding language in visual events \cite{rd_eval2}. However, many benchmarks suffer from the problem of \textit{visual content irrelevance}, which refers to models performing well on the benchmark datasets by relying primarily on the language modality \cite{rd_eval1}. Furthermore, data leakage and contamination problems \cite{rd_eval1} also hinder fair and accurate testing of a model's skills using benchmarks.

While it is important to continue directing research efforts towards developing more extensive multi-image benchmarks such as ReMI \cite{rd_eval3}, we argue that the purpose of evaluation is to also provide insights that can be directly leveraged for improving model architectures and learning procedures. For vision-to-language tasks involving single images, such as visual question answering and image captioning, recent work has used various interpretability methods for obtaining insights about relations between the visual and textual modalities of data \cite{VL-SHAP,MM-SHAP}, and for explaining model behavior \cite{rd_eval_BI2,rd_eval4,rd_eval_BI1}. There has also been work that focuses on understanding the functioning of general-purpose VLMs by analyzing the internal mechanisms and interactions between their components \cite{rd_eval5,rd_eval6}. These methods can complement traditional evaluation techniques for enabling intra- and inter-model comparisons to holistically understand behaviors and representations required for conditioning language in the visual data \cite{vl_mech_interp}. For scenarios dealing with language generation from visual events, interpretability methods are largely unexplored. We strongly advocate for efforts to develop interpretability techniques tailored for NLG tasks dealing with visual events, and also for adapting existing approaches from unimodal and vision-to-language settings involving static images.

\section{Conclusion}
In this paper, we focused on the problem of generating language from visual events and gathered, under the same umbrella, various tasks that are typically considered separate by the research community. Despite having different characteristics in terms of their communicative intents or input-output data, we argued that all the tasks present a common set of challenges for developing models and assessing their generated outputs. To understand the progress made over the years, we reviewed representative modeling approaches and evaluation protocols for each of the various exemplificative tasks we considered, and discussed them in a unified manner. We also presented a small-scale quantitative analysis in terms of the properties of datasets used to train models for the different tasks considered, regarding their visual and textual modalities. This type of data analysis can help to better understand the requirements of each task. The problem of generating text conditioned on sequences of multiple temporally ordered images or video frames has various real-world applications, which are becoming more present in our everyday lives. To facilitate further advancements, we underlined the challenges posed by this problem and proposed several concrete research directions informed by insights from linguistics, cognitive sciences, and natural language processing. We argue that leveraging these insights could also help the development of better VLMs, which are currently not immune from some of the highlighted limitations.

\begin{acks}
  We are grateful to our colleagues at the Dialogue Modelling Group (DMG) for their invaluable suggestions at different stages of this work. We thank Alberto Testoni and Anna Bavaresco for their insightful feedback. Raquel Fern{\'a}ndez was supported by the European Research Council (ERC Consolidator Grant DREAM 819455).
\end{acks}

\bibliographystyle{ACM-Reference-Format}
\bibliography{refs}

\end{document}